# Binarizing Business Card Images for Mobile Devices

A. F. Mollah[#], S. Basu[*], N. Das[*], R. Sarkar[*], M. Nasipuri[*], M. Kundu[*]

[#] *School of Mobile Computing and Communication, Jadavpur University, Kolkata, India*
`afmollah@gmail.com`

[*] *Department of Computer Science & Engineering, Jadavpur University, Kolkata, India*

*Abstract*— Business card images are of multiple natures as these often contain graphics, pictures and texts of various fonts and sizes both in background and foreground. So, the conventional binarization techniques designed for document images can not be directly applied on mobile devices. In this paper, we have presented a fast binarization technique for camera captured business card images. A card image is split into small blocks. Some of these blocks are classified as part of the background based on intensity variance. Then the non-text regions are eliminated and the text ones are skew corrected and binarized using a simple yet adaptive technique. Experiment shows that the technique is fast, efficient and applicable for the mobile devices.

## I. INTRODUCTION

With the pervasive availability of low cost portable imaging devices, digital camera has become so popular that majority of the mobile devices such as cell-phones and Portable Digital Assistants (PDA) have inbuilt digital camera. The resolution of these cameras is getting increased day by day. The computing power and primary memory of the mobile devices are also gradually going higher. So, the idea of image processing and analysis is no more limited to desktop computation. Researchers have paid significant attention towards developing Optical Character Recognition (OCR) systems for document images on mobile devices. High resolution flatbed scanners are used for desktop processing of document images. Scanner captured document images hardly suffer from blur, shadow, skew and perspective distortion whereas these are very common for camera captured documents. On the other hand, mobile devices are portable and so more useful than scanners for document processing, particularly for capturing and processing any arbitrary documents such as thick books, fragile documents like old historical manuscripts, scene texts, caption texts, graphic texts, etc.

Business Card Reader (BCR) for mobile devices is such a useful application of camera captured document image processing. With the development of an efficient BCR system, the information of the acquired business card images can be directly populated to the contact profile of the mobile devices. Thus, the business card management will be of great ease than ever before. None would have to carry the business card album nor have to type the information of the cards to populate into the handheld devices.

One of the major challenges of designing such a system is the binarization of business card images. Business card images often have complex background and texts of multiple natures. These images may contain logo, picture, texts of different fonts and various font sizes, graphic background, etc. Therefore, binarization can not be done straight forward and we found that neither global nor locally adaptive binarization techniques [1-4] can be applied for such images. Majority of the conventional binarization techniques are for scanned documents and some are for camera captured document images too.

Seeger et al. [5] has presented an algorithm based on Background Surface Thresholding (BST) for binarizing camera captured document images. At first, the texts are eliminated from the document image and then the background is interpolated. They compute a threshold with the help of this interpolated background image and the original image, and binarize the document using it.

Kim et al. [6] has presented an adaptive multi-window binarization method for document images acquired by camera. Information from the global trend as well as the local detail is used to determine the local threshold by applying multiple windows varying in sizes.

Stroke neighborhood enhancement based document image binarization method is found in [7]. The foreground pixels are marked and the strokes are enhanced based on their neighborhood information. Then the enhanced image is binarized by incorporating contrast enhancing function and the smooth function. Segmentation based binarization approach is found in [8]. Literatures suggest that most of the binarization methods are for document images. They may not be directly applicable for binarizing business card images.

In this paper, we have presented a novel technique for binarizing business card images, designed in our work towards developing an efficient BCR system for mobile devices. Experiments show that the technique has low computational overhead, and is fast and efficient.

## II. THE PRESENT METHOD

As discussed in Section I, binarization of business card images may not be straight forward. Our technique consists of few stages. At the first stage, the background is eliminated using a method as discussed in Section II-A. Then the foreground non-text elements are removed as explained in Section II-B. After that the image is expected to contain only text regions. These text regions are skew corrected as illustrated in Section II-C. And finally, the skew corrected test regions are binarized in Section II-D.

## A. Background Elimination

The entire image is at first divided into blocks of a fixed size. The more is the length of the block, the more is the number of horizontally contiguous words included in a single text region. And similarly, the less the height of the block is, the less is the possibility that a block covers more than one text lines. So, we have mostly experimented with rectangular blocks. The width and height is varied and tuned for best results and we found that it works well for the block of width $W/64$ ($W$ is the width of the card image) and height 2 pixels. Next, we classify each block as either an information block or a background block based on the intensity variance within it. An information block belongs to either a text region or an image region including noise. The motivation behind this approach is that the intensity variance is low in case of background blocks and high in case of information blocks. So, if the intensity variance of a block is lesser than a dynamic threshold ($T_\sigma$) as given in Eq. (1), it is considered as a background block. Otherwise, the block is considered as an information block. But, no block is classified as background until the minimum intensity within the block exceeds a heuristically chosen threshold ($T_{min}$). The formulation of $T_\sigma$ is described below.

$$T_\sigma = T_{fixed} + T_{var} \quad (1)$$
$$T_{var} = [(G_{min}-T_{min}) - min(T_{fixed}, G_{min}-T_{min})] * 2 \quad (2)$$

where, $G_{min}$ and $G_{max}$ are the minimum and maximum gray level intensity of the pixels in a block respectively and $T_{fixed}$ is the minimum intensity tolerance subject to tuning.

All the pixels of the blocks identified as background in this section are assigned the maximum intensity i.e. 255 to denote that they are part of the background.

## B. Removal of non-text regions

A connected component (CC) in the card image may represent a picture, logo, noise or a text region. We focus to identify only the text regions using rule-based classification. Whenever we identify a CC, we apply some heuristically chosen rules and decide whether it is a text region or not. We have played a little conservative in the approach so that we do not loose any text region, however unclear it is, even if we classify a non-text region as a text one.

A text region can not be too small because it must contain at least one character. Besides that, small characters are very unlikely to be isolated in the card. So, if the height and width of a CC are lesser than that of the smallest possible character of the card, then the region could be classified as noise.

Let, $H_{cc}$ and $W_{cc}$ be the height and width of a connected component respectively and $A_{cc}$ be its area in terms of pixels. Then, the CC is classified as noise if any of the following three conditions (Eq. (3-5)) gets satisfied. Here, $H_{TH}$, $W_{TH}$ and $A_{TH}$ denote minimum height, width and area of a CC respectively.

$$H_{cc} < H_{TH} \quad (3)$$
$$W_{cc} < W_{TH} \quad (4)$$
$$A_{cc} < A_{TH} \quad (5)$$

Some business cards may contain lines along either or both horizontal and vertical directions. We consider a CC to be a horizontal line if Eq. (6) is satisfied and a vertical line if Eq. (7) is satisfied. $L_{TH}$ and $B_{TH}$ denote the minimum length and maximum breadth of a line.

$$H_{cc} < B_{TH} \ \& \ W_{cc} > L_{TH} \quad (6)$$
$$W_{cc} < B_{TH} \ \& \ H_{cc} > L_{TH} \quad (7)$$

Typically, a text region has a certain range of width to height ratio ($R_{w2h}$). We consider a CC as a text if $R_{w2h}$ lies within the range ($R_{min}$, $R_{max}$).

$$R_{min} < R_{w2h} < R_{max} \quad (8)$$

To remove the logo(s), we have made an assumption that neither a horizontal nor a vertical line can be drawn through a logo and it is not as small as a large possible character within the card. Thus, logos and other noises satisfying the above assumption get eliminated.

Another important property of text region is that in a text region, the number of foreground pixels is significantly less than that of the background pixels. We consider a certain range of ratio of foreground pixels to the background ones ($RA_{cc}$) given by ($RA_{min}$, $RA_{max}$) for candidates of text region.

## C. Skew Correction

Skew angle is estimated for each text region and then the text region is rotated accordingly to get skew corrected. To calculate the skew angle, we consider the bottom profile of the gray shade of a text region. It may be noted that the gray shade is the background of the card around the text regions. The assumption is that the background of a camera captured card image will not be of the maximum intensity. The profile contains the heights in terms of pixel from the bottom edge of the bounding rectangle formed by the text region to the first gray/black pixel found while moving upward. However, if the extent of the gray shade along the column of a profile is too small, we discount it as a valid profile.

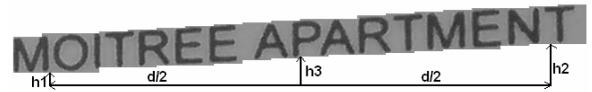

Fig. 1: Skew Angle Computation

As the profile is ready, we calculate the mean ($\mu$) and the mean deviation ($\tau$) of heights as shown in Eq. 9 and 10 respectively. The computation of mean deviation does not involve floating point arithmetic. Although, we can convert the floating point arithmetic to integer one, we want to avoid it as our intent is to embed the method on mobile devices that usually do not have a Floating Point Unit (FPU). Then, we exclude some elements of the profile that are not in sync with the others i.e. not within (+$\tau$, -$\tau$). These elements hardly contribute to the actual skew of the text region and so get eliminated.

$$\mu = \frac{1}{N}\sum_{i=0}^{N-1} h[i] \qquad (9)$$

$$\tau = \frac{1}{N}\sum_{i=0}^{N-1}|\mu - h[i]| \qquad (10)$$

where $N$ is the profile length, $h$ is the profile array and $h[i]$ denotes the height at $i^{th}$ position.

Among the remaining profile elements that really contribute to the actual skew of the text regions, we consider the leftmost ($h1$), rightmost ($h2$) and the middle profile element ($h3$) as shown in Fig. 1. The distance between $h1$ and $h2$ is computed as $d$. Then, the individual skew angles for the slope between $h1$ and $h2(\alpha)$, $h1$ and $h3(\beta)$, and $h2$ and $h3(\gamma)$ are computed as formulated in Eq. 11-13 respectively. Now, ideally they should be the same. We introduce a threshold ($\varepsilon$) to allow a certain deviation in between them. So, if none of the deviations between any two of $\alpha$, $\beta$ and $\gamma$ is more than $\varepsilon$, we take an average and rectify the skew of the text region. Otherwise, we look forward to the top profile of the text region and compute the skew angle. Respective skew angles as computed from the top profile of the text region are $\alpha'$, $\beta'$ and $\gamma'$. If these are found to be inline, we take an average of them and rectify the skew. Else, the smaller one between the averages obtained from top and bottom profiles is considered as the skew angle. It may be noted that this approach gives a mean to bypass some computation if not required.

$$\alpha = \arctan\left(\frac{\delta h}{d}\right), \ \delta h = h2 - h1 \qquad (11)$$

$$\beta = \arctan\left(\frac{\delta h}{d}\right), \ \delta h = h3 - h1 \qquad (12)$$

$$\gamma = \arctan\left(\frac{\delta h}{d}\right), \ \delta h = h2 - h3 \qquad (13)$$

*D. Binarization*

As a CC is classified as a text region, it is binarized with adaptive yet simple technique. If the intensity of a pixel within the CC is less than the mean of the maximum and minimum intensities of a CC, it is taken as a foreground pixel. Otherwise, we check the 8 neighbors of the pixel and if any 5 or more neighbors are foreground, then also we consider the pixel as a foreground one. It may be noted that the border pixels do not have 8 neighbors and so will not be subject to this technique. The remaining pixels are considered as part of the background. The algorithm is given in Table I.

The advantage of this approach of binarization is that the disconnected foreground pixels of a character are likely to become connected due to neighborhood consideration. Instead of having efficient binarization techniques, we have designed this simple algorithm keeping the computational constraints of the mobile devices in view.

TABLE I
BINARIZATION ALGORITHM

```
for all pixels (x, y) in a CC
    if Intensity(x, y) < (G_min + G_max)/2, then
        mark (x, y) as foreground
    else
        if no. of foreground neighbors > 4, then
            mark (x, y) as foreground
        else
            mark (x, y) as background
        end if
    end if
end for
```

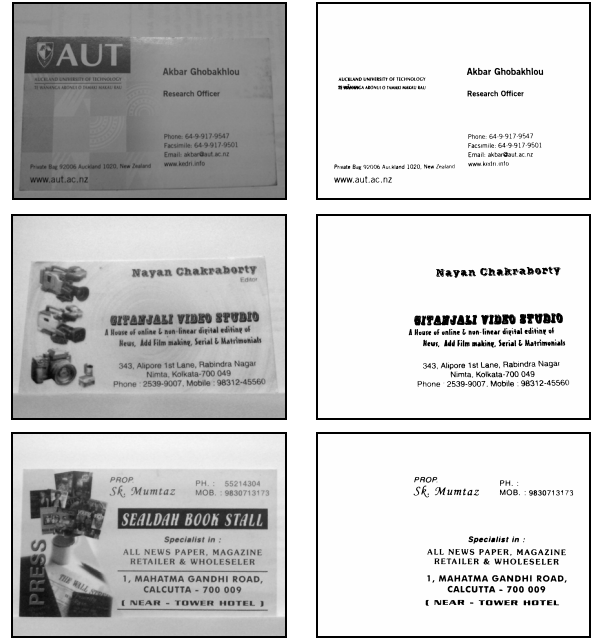

Fig. 2: Sample cards (left) and their skew corrected binarized view (right)

III. EXPERIMENTAL RESULTS AND DISCUSSION

We have experimented on a dataset of 100 business card images of various types acquired with a cell-phone camera (Sony Ericsson K810i) to evaluate the performance of the present binarization technique. The dataset consists of both simple and complex cards containing complex backgrounds and logos. Some cards contain multiple logos and some logos are combination of text and image. Most of the images are skewed, perspectively distorted and degraded.

*A. Efficiency of the Technique*

One of the sole purposes of binarizing business card images is elimination of the non-text regions. So, we have estimated the efficiency of the present technique in terms of text region extraction accuracy. Once a text region is identified, it can be binarized using any binarization technique. Keeping the computational constraints of the mobile devices in view, we have implemented our own technique as discussed in Section II-D which is computationally efficient. To quantify the text extraction accuracy, we have designed the following method.

A text or a background region can be identified either way. Background here refers to all non-text regions. So, there can be four possible cases as shown in Table II. BB and TT are considered as true classifications whereas BT and TB are false ones.

TABLE II
JUSTIFICATION OF CC CLASSIFICATION

| CC/Region | Classified as | Justification |
|---|---|---|
| Background | Background | BB (True) |
| Background | Text | BT (False) |
| Text | Background | TB (False) |
| Text | Text | TT (True) |

So, the text extraction accuracy of the current technique is defined as

$$\text{Accuracy} = \frac{\text{No. of true classifications}}{\text{Total no. of CCs}} \quad (14)$$

Following the above text extraction accuracy definition, we have got a maximum mean accuracy of 98.93% for 3 Mega pixel images with $T_{fixed}$ = 20, $T_{min}$ = 100, $H_{TH}$ = H/60, $W_{TH}$ = W/40, $A_{TH}$ = W*H/1500, $B_{TH}$ = H/100, $L_{TH}$ = W/40, $R_{min}$ = 1.2, $R_{max}$ = 32, $RA_{min}$ = 5 and $RA_{max}$ = 90. Table III shows the accuracy rates when experimented with other resolutions on a moderately powerful desktop (PIV 2.4 GHz Processor, 256 MB RAM, 1 MB L2 Cache).

TABLE III
CLASSIFICATION ACCURACY WITH VARIOUS RESOLUTIONS

| Resolution (width x height) | Mean Accuracy (%) |
|---|---|
| 640x480 (0.3 MP) | 97.80 |
| 800x600 (0.45 MP) | 97.86 |
| **1024x768 (0.75 MP)** | **98.54** |
| 1182x886 (1 MP) | 98.00 |
| 1672x1254 (2 MP) | 98.45 |
| 2048x1536 (3 MP) | 98.93 |

*B. Applicability on Mobile Devices*

The applicability of the presented technique on mobile devices is checked by its computational requirements. As our aim is to deploy the proposed method into mobile devices, we want to develop a light-weight Business Card Reader (BCR) system beforehand and then to embed into the devices.

An observation [9] reveals that the majority of the processing time of a camera based optical character recognition (OCR) engine embedded into a mobile device is consumed in preprocessing including binarization. Although, we have shown the computational time of the presented method with respect to a desktop, the total time required to run the developed method on mobile devices will be tolerable. Fig. 3 shows the computation time with various resolutions.

As, limited memory is another constraint of the mobile devices, the presented method is designed to work with low memory requirement. Memory consumption is approximately 2-3 folds of the input image size.

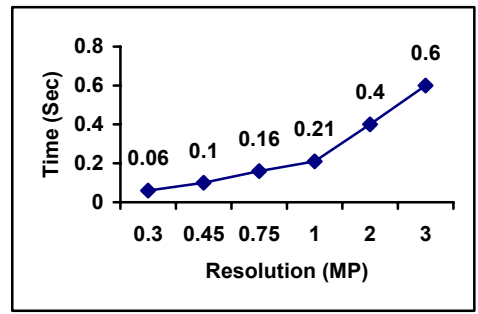

Fig. 3: Computation Time with Various Resolutions

*C. Discussion*

Although, one can see that the developed method works well as shown in Fig. 2, it too has certain limitations. Sometimes, the dot of 'i' or 'j' gets removed. When text and image/logo are very near to each other, they together form a single CC and get wrongly classified as background or text as shown in Fig 4. The current technique does not work for the reverse texts i.e. light texts on dark background as shown in the last image of Fig. 2. All these shortcomings of the present system will be taken up in our future work.

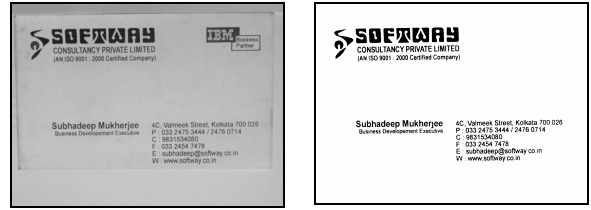

Fig. 4: Sample card (left) in which the logo is misclassified as text (right)

IV. CONCLUSIONS

We have presented and evaluated a method of binarizing mobile camera captured business card images and our experiments show that the result is satisfactory. It has been observed from this experimentation, that with the increase in image resolution, the computational time and memory requirements increase proportionately. Although, the maximum text region isolation accuracy is obtained with 3 mega pixel resolution, it involves high memory requirement and 0.6 seconds of processing time. It is evident from the findings that the optimum performance is achieved at 1024x768 (0.75 MP) pixels resolution with a reasonable accuracy of 98.54% and significantly low (in comparison to 3 MP) processing time of 0.16 seconds.

As we have employed a little conservative approach and got some background CCs as text ones (BT), we hope that these will get removed in segmentation and recognition. The occurrence of text classified as background (TB) is absolutely less which reveals good potential for the developed method discussed in this paper. Moreover, the presented method is directly applied on the raw images which are heavily distorted. If the images are rectified and then passed to our engine, we hope and foresee a better accuracy.

ACKNOWLEDGMENT

Authors are thankful to the *Center for Microprocessor Application for Training Education and Research (CMATER)* and project on *Storage Retrieval and Understanding of Video for Multimedia (SRUVM)* of the Department of Computer Science and Engineering, Jadavpur University for providing infrastructural support for the research work. We are also thankful to the *School of Mobile Computing and Communication (SMCC)* for proving the research fellowship to the first author.

REFERENCES

[1] N. Otsu, A threshold selection method from gray level histogram, IEEE Transaction SMC-9, pp. 62-66, 1979
[2] W. Niblack, An Introduction to Image Processing, Prentice-Hall, Englewood Cliffs, NJ, 1986, pp. 115-116
[3] O. D. Trier, Goal-directed evaluation of binarization methods, IEEE PAMI Vol. 17, No. 12, 1995
[4] J. Sauvola and M. Pietikainen, Adaptive document image binarization, Pattern Recognition, Vol. 33, pp. 225-236, 2000
[5] M. Seeger and C. Dance, Binarizing camera images for OCR, Proceeding 6th ICDAR, 54-58, 2001.
[6] I. J. Kim, Multi-window binarization of camera image for document recognition, Proc. of Int'l Workshop on Frontiers in Handwriting Recognition, pp. 323–327, 2004.
[7] Y. Zhu, C. Wang and R. Dai, Document Image Binarization Based on Stroke Enhancement, Proc. of 18th International Conference on Pattern Recognition (ICPR'06) Volume 1, pp. 955-958, 2006
[8] C. Thillou and B. Gosselin, Segmentation-based Binarization for Color Degraded Images, Proc. of International Conference on Computer Vision and Graphics (ICCVG 2004), Warsaw (Poland), pp. 808-813, 2004
[9] M. Laine and O. S. Nevalainen, A Standalone OCR System for Mobile Cameraphones, 17th Annual IEEE International Symposium on Personal, Indoor and Mobile Radio Communications, Sept. 2006, pp. 1-5.